%% file: main.tex
\documentclass[a4paper,conference]{IEEEtran}

\usepackage{cite}

\ifCLASSINFOpdf
  \usepackage[pdftex]{graphicx}
\else
\fi
\ifCLASSOPTIONcompsoc
 \usepackage[caption=false,font=normalsize,labelfont=sf,textfont=sf]{subfig}
\else
 \usepackage[caption=false,font=footnotesize]{subfig}
\fi
\usepackage{url}

\usepackage[english]{babel}

\usepackage{xcolor}

\usepackage{soul}

\usepackage[range-phrase={\,--\,}, range-units=single, per-mode=symbol, detect-all]{siunitx}

\makeatletter
  \@addtoreset{paragraph}{section}
\makeatother

\usepackage[acronym]{glossaries}
\newacronym{qos}{QoS}{Quality of Service}
\newacronym{qoe}{QoE}{Quality of Experience}
\newacronym{ai}{AI}{artificial intelligence}
\newacronym{ml}{ML}{machine learning}
\newacronym{fl}{FL}{federated learning}
\newacronym{nn}{NN}{neural network}
\newacronym{rem}{REM}{radio environment map}
\newacronym{drl}{DRL}{deep reinforcement learning}
\newacronym{sdr}{SDR}{software-defined radio}
\newacronym[firstplural=radio access technologies]{rat}{RAT}{radio access technology}
\newacronym{ran}{RAN}{radio access network}
\newacronym{tcas}{TCAS}{traffic alert and collision avoidance system}
\newacronym{ga}{GA}{general aviation}
\newacronym{mcs}{MCS}{Modulation and Coding Scheme}
\newacronym{fec}{FEC}{forward error correction}

\usepackage{balance}
\usepackage{cleveref}
\usepackage[all]{nowidow}
\usepackage{url}

\begin{document}

\setlength{\topmargin}{-0.698in}
\addtolength{\textheight}{-0.059in}
%

\title{Communication-Efficient Federated Learning for Tiny Language Models Predicting Features in Mobile Network Data}
\title{Efficient Federated Learning Tiny Language Models for Mobile Network Feature Prediction}



%
\author{
    \IEEEauthorblockN{
        Daniel Becking\IEEEauthorrefmark{3},
        Ingo Friese\IEEEauthorrefmark{4},
        Karsten Müller\IEEEauthorrefmark{3},
        Thomas Buchholz\IEEEauthorrefmark{4},\\
        Mandy Galkow-Schneider\IEEEauthorrefmark{4},
        Wojciech Samek\IEEEauthorrefmark{3},
        Detlev Marpe\IEEEauthorrefmark{3}
    }
        \IEEEauthorblockA{
        \IEEEauthorrefmark{3}
        Fraunhofer Heinrich Hertz Institute (HHI), Berlin, Germany, 
        \IEEEauthorrefmark{4}
        Deutsche Telekom AG, Berlin, Germany 
    }
        \IEEEauthorblockA{
        Email: \{daniel.becking, karsten.mueller\}@hhi.fraunhofer.de,\\ \{ingo.friese, thomas.buchholz, mandy.galkow-schneider\}@telekom.de
    }

}

%
\maketitle


\begin{abstract}

In telecommunications, Autonomous Networks (ANs) automatically adjust configurations based on specific requirements (e.g., bandwidth) and available resources. These networks rely on continuous monitoring and intelligent mechanisms for self-optimization, self-repair, and self-protection, nowadays enhanced by Neural Networks (NNs) to enable predictive modeling and pattern recognition.
Here, Federated Learning (FL) allows multiple AN cells --- each equipped with NNs --- to collaboratively train models while preserving data privacy. However, FL requires frequent transmission of large neural data and thus an efficient, standardized compression strategy for reliable communication.
To address this, we investigate \textit{NNCodec}, a Fraunhofer implementation of the ISO/IEC Neural Network Coding (NNC) standard, within a novel FL framework that integrates tiny language models (TLMs) for various mobile network feature prediction (e.g., ping, SNR or band frequency). 
Our experimental results on the \textit{Berlin V2X} dataset demonstrate that NNCodec achieves transparent compression (i.e., negligible performance loss) while reducing communication overhead to below 1\%, showing the effectiveness of combining NNC with FL in collaboratively learned autonomous mobile networks.
\end{abstract}


\section{Introduction}

\begin{figure}[b]
    \centering
    \includegraphics[width=.724\linewidth]{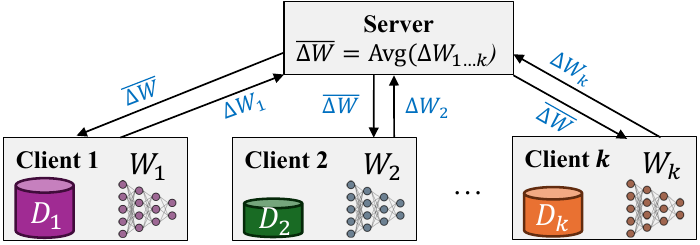}
    \caption{Federated averaging differential weight updates $\Delta W_i$.}
    \label{fig:fl}
\end{figure}

With the advancement of 5G, 6G, and beyond, assessing Quality of Service (QoS) and user Experience (QoE) is essential. Wireless communications are expected to integrate advanced machine learning technologies, and one promising strategy is to equip mobile network cells with NNs that collaboratively learn in an FL setup, enhancing overall QoS and QoE.
In FL, a shared NN is trained across multiple client devices, each utilizing its own data subset (see Figure \ref{fig:fl}). A central server coordinates by aggregating updates from clients and redistributing them. While FL offers data protection by keeping data local, it faces challenges like communication overhead due to frequent NN data exchanges, often constrained by bandwidth. 
To address this, effective data compression is crucial to minimize communication overhead and latency. Two main strategies are: (i) reducing the frequency of weight updates (communication delay) by performing multiple local train iterations before transmitting updates, and (ii) compressing transmitted data through parameter reduction (e.g., pruning, sparsification, or decomposition), precision reduction (quantization), and lossless compression (e.g., entropy coding or sparse matrix formats).

In this work, we employ FL to train tiny language models on cellular data from the Berlin V2X dataset~\cite{hernangomez_berlin_2023} to predict quality features such as \texttt{ping}, \texttt{RSSI},  and \texttt{jitter}, based on input features like \texttt{MCS} (Modulation and Coding Scheme), \texttt{Tx\_Power}, \texttt{frequency}, GPS features and environmental side information. The neural data communicated between clients and server is compressed using technology from the ISO/IEC Neural Network Coding (NNC) standard \cite{becking_neural_2024}.

\section{Methodology}

\subsection{Tiny Language Models (TLMs) for Cellular Data}

Transformer-based language models demonstrate remarkable potential in capturing complex correlations within data through the attention mechanism. They effectively identify relationships between so-called tokens in the input sequence, which enables the model to comprehend context. This comprehension allows the model to generate relevant responses by predicting the probability distribution across its entire vocabulary (i.e., set of unique tokens the model can handle) and determining the most probable next token.

\begin{figure}
    \centering
    \includegraphics[width=1\linewidth]{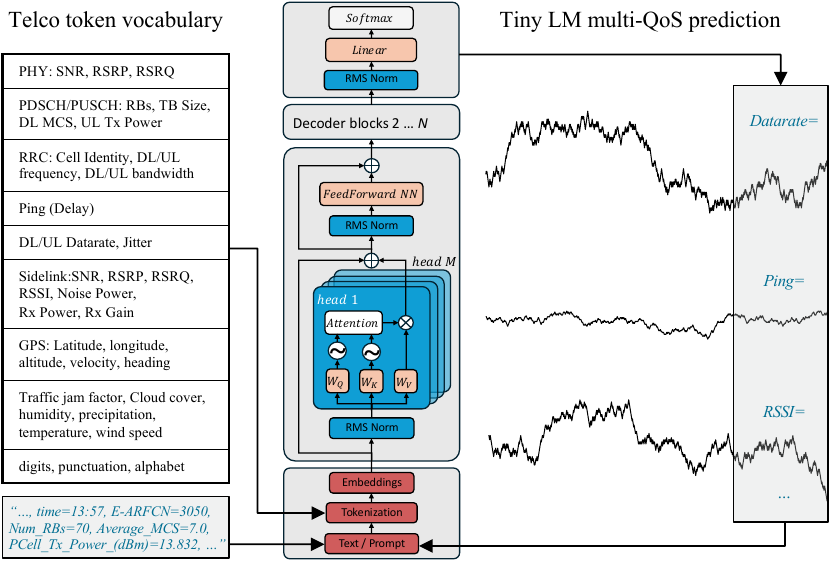}
    \caption{Concept scheme of TLM-based cellular feature predictions.}
    \label{fig:overview}
    \vspace{-2mm}
\end{figure}

As illustrated in Figure \ref{fig:overview}, our work studies the technical feasibility of such models to capture correlations in cellular data by converting from a tabular format into strings and subsequently into tokens. Then, each input sequence represents a concatenated set of cellular features as collected in the Berlin V2X dataset.
We utilized a custom tokenizer and lightweight version of Llama 2 \cite{touvron_llama_2023}, built on the 
\textit{llama.c}~\cite{karpathy_llama2c_nodate} implementation. 
We deployed three model sizes, with parameter counts of 409k, 10M, and 32M, corresponding to memory footprints of 1.6MB, 40MB, and 127MB, respectively.

\subsection{Berlin V2X Data Preprocessing and Tokenization}


The Berlin V2X dataset  \cite{hernangomez_berlin_2023} provides high-resolution, GPS-located wireless measurements in various urban settings along a 17.2 km long route through Berlin city. Data collection involved up to four cars driving the route 17 times over a span of three days. The dataset contains 159 features including information on physical layer parameters, cellular radio resource management, QoS parameters, GPS positioning data, and additional context like traffic and weather conditions.

For our analysis, we filter data points that include operator~1 (i.e., connected to Deutsche Telekom’s network), resulting in 103,554 GPS-located data points. From this selection, we randomly picked 10\% for testing. The remaining data is split according to the associated area segments 'Residential', 'Park', 'Avenue', 'Highway', 'Tunnel', which results in 28,017, 33,487, 19,747, 10,390 and 475 training samples for the five respective AN clients participating in FL.

For training language models, the data is usually tokenized, meaning that the input is divided into units called tokens. We configured a custom SentencePiece \cite{kudo_sentencepiece_2018} tokenizer specifically designed for the vocabulary needed to capture the cellular communication of the Berlin V2X dataset.
Our initial, naïve approach involves predicting numbers token-by-token. For example, the sequence \texttt{PCell\_RSRQ\_max = -7.445} is broken down into the tokens: \texttt{PCell\_RSRQ\_max}, \texttt{=}, \texttt{-}, \texttt{7}, \texttt{.}, \texttt{4}, \texttt{4}, \texttt{5}, and the whitespace characters.

\subsection{Efficient FL between TLM-equipped Network Cells}
Efficient FL requires a suitable combination of model complexity, data compression and local learning.
We integrated NNCodec~\cite{becking_nncodec_2023}, an open-source implementation of the NNC standard (ISO/IEC 15938-17), to compress communication among five TLMs engaged in an FL task predicting cellular network features.
For encoding the differential updates $\Delta W_i$ between $i\in\{1,..,5\}$ clients and the server, we use NNC's encoder-decoder architecture as detailed in~\cite{becking_neural_2024}. 
The process begins with parameter reduction techniques such as sparsification and pruning, enhancing efficiency by zeroing out many values. Quantization follows, controlled by a quantization parameter ($qp$), and concludes with optimized arithmetic coding using \textit{DeepCABAC}.
The NNC standard also provides syntax for coordinating decoded incremental weight updates in distributed learning. The Parameter Update Tree (PUT)~\cite{becking_neural_2024} offers a referencing mechanism to uniquely identify coded entities relative to previous versions, supporting flexible communication, partial NN updates, (a)synchronous settings, client failures, and new client integration.

\vspace{1.5mm}
\section{Experimental Results}



\begin{figure}[!t]
    \centering
    \includegraphics[width=1.0\linewidth]{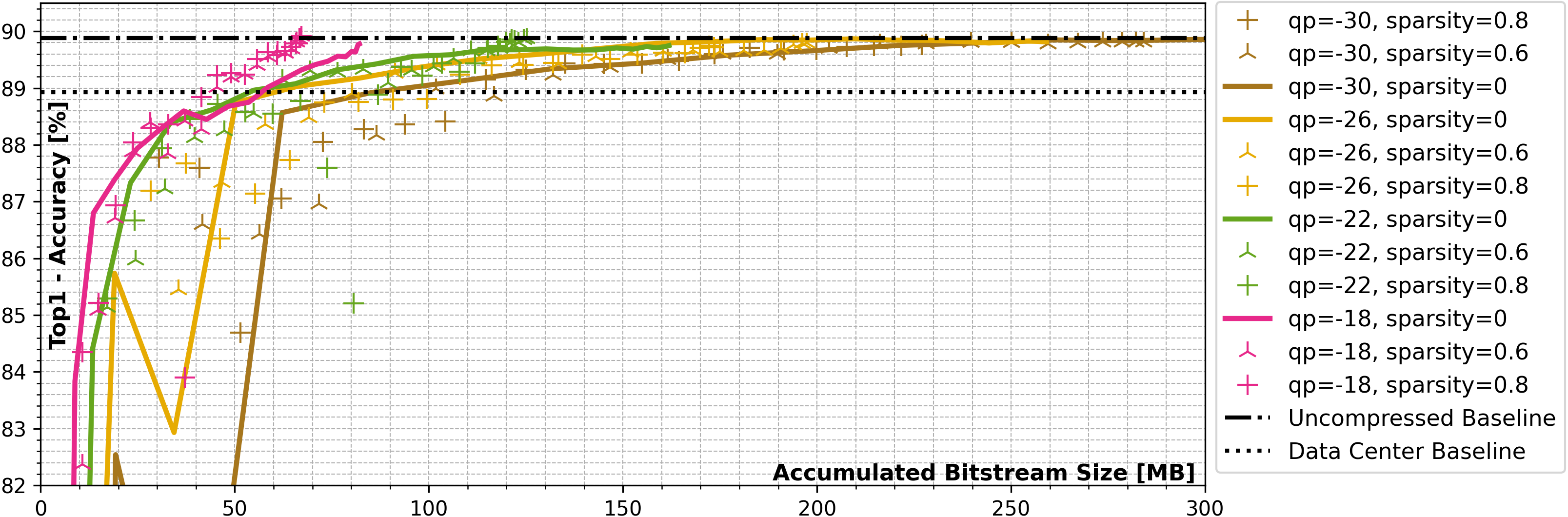}
    \caption{Coding performance of various $qp$ and sparsity values with a 
    fixed TLM\_size of 1, learning rate 3e-4, y-axis limited to values $\geq$82\%.}
    \label{fig:results}
    \vspace{-1.5mm}
\end{figure}

All experiments ran for 25 FL communication rounds. In the uncompressed baseline setting, TLMs achieved over 89\% $\text{top-1}$ accuracy for next-token prediction and perplexities below 1.35, demonstrating their ability to capture correlations in cellular data and predict missing features. Various compressed settings are shown in Figure~\ref{fig:results}.
Table \ref{tab:coding_results} details selected runs and compression efficiency. For TLM\_size 1 and 2, one selected case achieved transparency with minimal communicated data, while another surpassed baselines at the cost of additional data. For TLM\_size 0, only three runs achieved transparent compression. 
In conclusion, compression rates below 1\% of the original size with negligible performance degradation are feasible in the presented use case.


\input{coding_res_table}






\vspace{1mm}
\bibliographystyle{IEEEtran}
\bibliography{bibliography.bib}


\end{document}

%% file: coding_res_table.tex
\begin{table}[t]
    \centering
    \small
    \caption{Selected transparent coding results for different TLM sizes.}
    \resizebox{\columnwidth}{!}{
    \renewcommand{\arraystretch}{1.2} 
    \begin{tabular}{c c c c c c c c c c c}
        \hline
        \textbf{Size} & \textbf{lr} & $\mathbf{qp}$ & \textbf{Sparsity} & \multicolumn{2}{c}{\textbf{Performance}} & \multicolumn{2}{c}{\textbf{Data [MB]}} & \textbf{Comp.} & \multicolumn{2}{c}{\textbf{Perf. Degradation}} \\
        & & & & Acc [\%] & PPL & Uncomp. & Comp. & \textbf{Ratio [\%]} & Acc [\%] & PPL \\
        \hline
        0 & 3e-4 & -26 & 0\%  & 89.35 & 1.35 & 424.64 & 12.84 & 3.02 & -0.18 & +0.01 \\
        1 & 5e-5 & -22 & 0\%  & 89.70 & 1.35 & 10,095.84 & 25.10 & 0.25 & -0.33 & +0.00 \\
        1 & 3e-4 & -18 & 80\% & 89.88 & 1.34 & 10,095.84 & 67.16 & 0.67 & +0.03 & -0.00 \\
        2 & 5e-5 & -22 & 60\% & 89.73 & 1.35 & 31,874.56 & 49.95 & 0.16 & -0.34 & +0.00 \\
        2 & 3e-4 & -22 & 0\%  & 89.70 & 1.35 & 31,874.56 & 487.12 & 1.52 & +0.11 & -0.00 \\
        \hline
        \vspace{-5mm}
    \end{tabular}
    }
    \label{tab:coding_results}
\end{table}